\documentclass[11pt]{article}

\usepackage{acl}
\usepackage{times}
\usepackage{latexsym}

\usepackage[T1]{fontenc}

\usepackage[utf8]{inputenc}

\usepackage{microtype}

\usepackage{inconsolata}

\usepackage{graphicx}

\usepackage{xcolor}         
\usepackage{booktabs}
\usepackage{graphicx}
\usepackage[table]{xcolor}
\usepackage{array}
\usepackage{enumitem}
\usepackage{multirow}
\newcolumntype{g}{>{\columncolor[HTML]{EAF0F6}}c}

\usepackage{tcolorbox}
\tcbuselibrary{listings, skins, breakable}
\usepackage{listings}
\usepackage{tabularx}
\usepackage{multirow}
\usepackage{array}
\usepackage{amsmath}
\usepackage{amssymb}
\usepackage{soul}
\usepackage{float}
\newcommand{\methodAbbre}{MERIT}

\title{MERIT: Matching Expertise via Rubric-Informed Training for Reviewer Assignment}

\author{
  Zixuan Yang\quad
  Yibo Zhao \quad
  Weicong Liu\quad
  Xiang Li\thanks{Corresponding Author: \texttt{xiangli@dase.ecnu.edu.cn}} \\
  School of Data Science and Engineering, East China Normal University
}

\begin{document}
\maketitle

\begin{abstract}
Matching submissions with suitable reviewers at scale is a growing challenge for major venues, yet existing approaches either rely on coarse proxy signals that conflate general relatedness with true suitability, or require expensive human annotations that are difficult to scale for training. We propose MERIT, a two-stage framework that bridges this gap by converting criterion-level expertise matching into scalable suitability supervision. In the first stage, we train a reviewer assessor via reinforcement learning to identify the expertise dimensions a paper requires, match them against the reviewer's prior work, and produce a suitability decision, with rewards provided by an LLM judge guided by paper-specific expertise rubrics. In the second stage, we distill the assessor's predictions into an embedding-based retriever for efficient large-scale assignment. Experiments show that our 4B reviewer assessor outperforms larger general-purpose LLMs on suitability classification, and the resulting retriever achieves state-of-the-art performance across LR-Bench and the CMU Gold dataset. Our code is available at \url{https://github.com/Luli3220/MERIT}.
\end{abstract}

\section{Introduction}\label{sec:introduction}

The quality of peer review hinges on matching each submission with reviewers who have the right expertise~\citep{HighQualityPR,expert}.
As submission volumes grow---major artificial intelligence venues now receive over ten thousand papers per cycle~\citep{PRProblem,PRProblem1}---manually identifying suitable reviewers has become infeasible, making automatic reviewer assignment an increasingly critical problem.

\begin{figure}[t]
    \centering
    \includegraphics[width=\linewidth,
        clip]{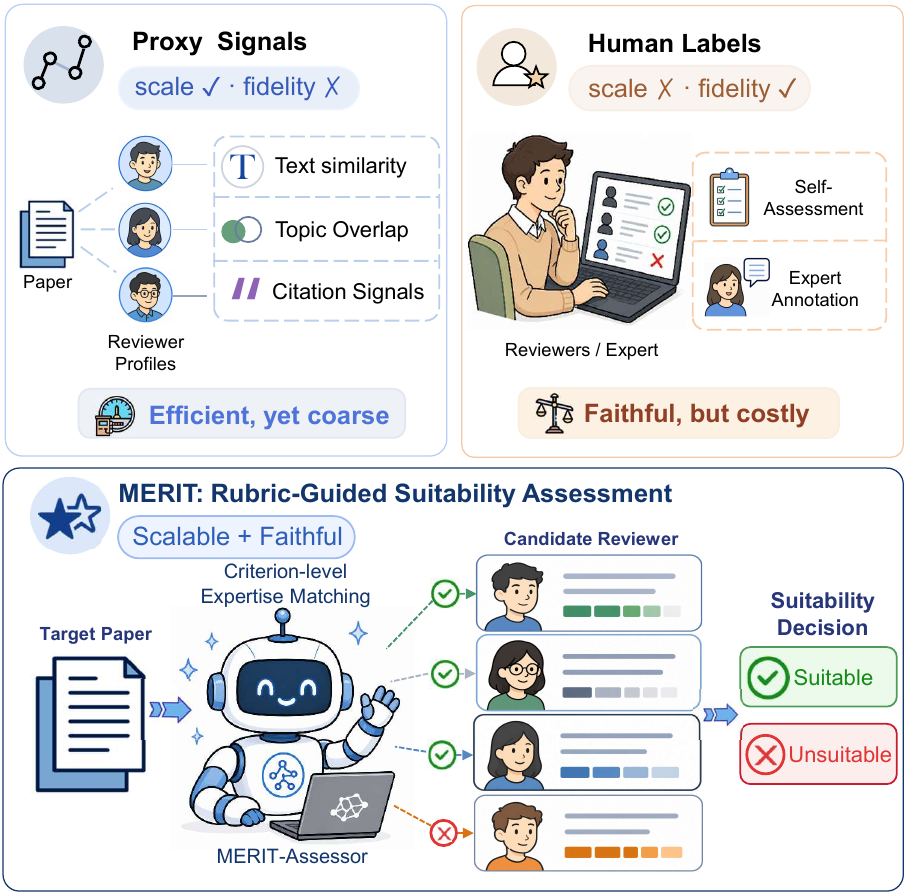}
\caption{Existing reviewer assignment approaches rely on either scalable but coarse proxy signals (left) or faithful but costly human labels (right). \methodAbbre{} (bottom) trains a reviewer assessor with paper-specific expertise rubrics for criterion-level suitability matching.}
    \label{fig:intro}
\end{figure}

Most existing methods estimate paper--reviewer affinity from scalable but coarse proxy signals: textual similarity~\citep{tpms,wordembedding}, topical overlap~\citep{topic1,topic2}, citation relations~\citep{SIGIR,tcrrec}, or combinations
thereof~\citep{CoF}. These signals collapse all overlap between a reviewer's prior work and the target paper into a single relevance score, treating every shared element as equally important regardless of its role in the paper. 
For instance, consider a paper whose core contribution is a novel policy-optimization algorithm evaluated on instruction-following benchmarks. A benchmark specialist may overlap with the paper's evaluation setting, while an optimization expert may overlap with its algorithmic contribution; an aggregate affinity score may fail to distinguish these roles, even though the latter expertise is more central to evaluating the paper's contribution.
The root issue is that a reviewer is truly \emph{suitable} only
when their expertise covers the dimensions most critical to the paper, yet proxy signals provide no paper-specific mechanism for distinguishing critical expertise from incidental overlap.

An alternative is to use higher-fidelity signals such as reviewer self-assessments~\citep{CMU,RATE} or expert annotations~\citep{CoF,NIPS}. 
These signals more directly capture whether a reviewer is suitable for a given paper, but they are expensive to collect and difficult to scale.
This creates a gap: \emph{scalable signals lack the granularity to capture suitability, while high-fidelity signals cannot be obtained at the scale required for training.}



Recent work has explored using large language models (LLMs) to assess reviewer suitability directly, but general-purpose LLMs remain unreliable for this task~\citep{CMU,RATE}, tending to produce broad judgments of topical relevance rather
than evaluating expertise along individual dimensions.
This may stem from the absence of an explicit specification of what expertise a paper demands: each paper requires expertise along a distinct set of dimensions, some critical to evaluating its core contribution and others secondary.
Without such a specification, LLMs have no basis for distinguishing central expertise from incidental overlap. 
We address this by decomposing each paper's expertise requirements into a set of weighted criteria---what we call a \emph{paper-specific expertise rubric}. 
Each criterion specifies one expertise dimension and is assigned an importance weight reflecting how critical it is to the paper.


Building on this formulation, we propose \methodAbbre{}, a two-stage framework. 
In the first stage, we train a reviewer assessor with reinforcement learning.
Given a paper and a reviewer's publication history, the model identifies the expertise dimensions the paper requires, evaluates whether the reviewer's publications demonstrate expertise in each dimension, and produces a suitability decision.
The reward signal is provided by an LLM judge that evaluates each model output against an automatically generated expertise rubric for the target paper, along two axes: whether the output addresses each criterion with grounded evidence (\emph{rubric coverage}), and whether the final decision is consistent with the evidence (\emph{decision consistency}).

Applying the reviewer assessor to every candidate reviewer is prohibitively costly at conference scale.
In the second stage, we therefore use the trained assessor to annotate a large set of candidate pairs and distill the supervision into an embedding-based retriever for efficient large-scale assignment.

In summary, our contributions are as follows:
\begin{itemize}[leftmargin=*, itemsep=0pt]
\item 
We formulate reviewer assignment as criterion-level expertise matching and introduce paper-specific expertise rubrics as a supervision signal that combines the scalability of proxy signals with the granularity of expert annotations.

\item
We propose~\methodAbbre{}, a two-stage framework: (i) training a reviewer assessor with rubric-guided reinforcement learning, and (ii) distilling its predictions into an embedding-based retriever for efficient large-scale assignment.

\item 
Experiments show that our 4B reviewer assessor outperforms larger general-purpose LLMs on reviewer assessment, and that the resulting retriever achieves state-of-the-art performance across LR-Bench and the CMU Gold dataset.
\end{itemize}

\section{Related work}\label{sec:relatedwork}


\paragraph{Automatic Reviewer Assignment.}
The quality of reviewer assignment depends largely on paper--reviewer affinity scores~\citep{scoreimportant1,scoreimportant2}, and prior work has explored a range of signals for estimating these scores. One line of work relies on scalable proxy signals. Early methods compute affinity from TF--IDF similarity~\citep{tpms} or latent topic models~\citep{topical1,topical2}
between papers and reviewer publications. More recent methods use
citation-informed document encoders such as SPECTER~\citep{specter} and SciNCL~\citep{scincl} to capture
richer semantic relatedness, while CoF~\citep{CoF} combines semantic, topical, and citation signals into a unified scoring framework. RATE~\citep{RATE} uses proxy-derived scores to construct weak supervision for training a dedicated retrieval model. Despite their scalability, these methods estimate affinity from aggregate paper--reviewer overlap without distinguishing which
expertise dimensions are most relevant to a given paper.

Another line of work obtains supervision from reviewer
self-assessments~\citep{CMU,RATE} or expert annotations~\citep{NIPS,CoF}. These signals more directly reflect suitability but demand substantial human effort, limiting their use as training data. 
Our work bridges these two lines by using LLMs to perform criterion-level expertise matching, producing supervision that captures fine-grained suitability while remaining scalable.


\paragraph{Rubric-based Rewards.}
Rubric-based evaluation decomposes the assessment of complex outputs into criterion-level judgments, and has been used to evaluate LLM outputs in open-ended generation~\citep{Geval,prometheus}, instruction
following~\citep{flask,infobench}, and domain-specific tasks~\citep{HealthBench,HealthSCORETS,PLawBench}. More recently, such criterion-level judgments have been aggregated into reward signals for LLM post-training~\citep{RaR,AdvancedIF,openrubrics}. 
In all these settings, rubrics define what constitutes a good \emph{output}. We apply the same rubric-based framework to a different object of evaluation: our rubrics define what constitutes a suitable \emph{reviewer}, specifying the expertise dimensions a paper requires and their relative importance.

\begin{figure*}[t]
    \centering
    \includegraphics[width=\linewidth]{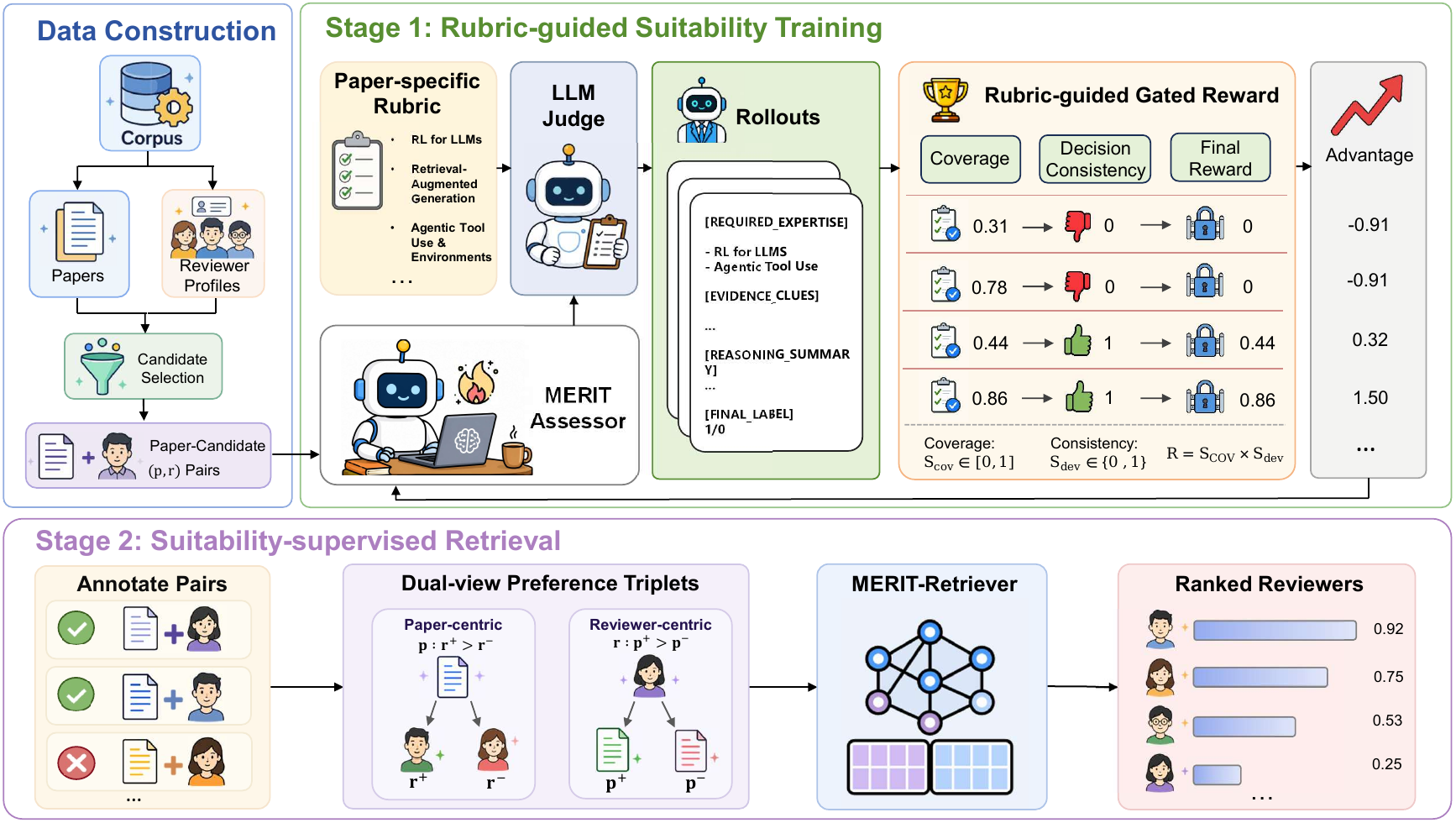}
    \caption{Overview of the full pipeline of our proposed method.}
    \label{fig:framework}
\end{figure*}

\section{Methodology}\label{sec:Methodology}

Given a target paper $p$ and a pool of candidate reviewers $\mathcal{R}$, where each reviewer $r \in \mathcal{R}$ is represented by their publication history $H_r = \{h_1, \ldots, h_{|H_r|}\}$, our goal is to identify reviewers whose expertise covers the dimensions most critical to evaluating $p$, efficiently and at scale. We propose \methodAbbre{}, a two-stage framework for \ul{M}atching \ul{E}xpertise via \ul{R}ubric-\ul{I}nformed \ul{T}raining for reviewer assignment, as illustrated in Figure~\ref{fig:framework}.

In the first stage, we train a reviewer assessor with reinforcement learning to perform criterion-level expertise matching and produce suitability decisions, with rewards provided by an LLM judge guided by automatically generated, paper-specific expertise rubrics (Section~\ref{sec:stage_1}).
In the second stage, we distill the trained assessor's predictions into an embedding-based retriever for efficient large-scale assignment (Section~\ref{sec:stage_2}).
We refer to the trained assessor and retriever as the \methodAbbre{}-Assessor and \methodAbbre{}-Retriever, respectively.
Both stages operate on paper–reviewer pairs constructed via a shared candidate retrieval procedure (Section~\ref{sec:DataConstruction}).


\subsection{Data Construction}
\label{sec:DataConstruction}

Training on randomly sampled paper--reviewer pairs would yield mostly trivial negatives with little topical overlap, offering limited signal for learning fine-grained suitability distinctions. 
We instead retrieve candidates using an existing affinity model (RATE; \citealp{RATE}), so that candidates have relatively high topical overlap with the target paper.
This makes the resulting pairs challenging: candidates are topically related to the target paper but vary in whether they are truly suitable to review it.

For each target paper, we augment its title and abstract with the introduction section to provide richer context for both rubric generation and suitability assessment.
We then retrieve the top-ranked reviewers as candidates, yielding paper--reviewer pairs that are split into two disjoint subsets: one for training the \methodAbbre{}-Assessor
(Section~\ref{sec:stage_1}) and another for generating 
pseudo-labels for the \methodAbbre{}-Retriever (Section~\ref{sec:stage_2}).
Detailed construction procedures, including filtering criteria and dataset 
statistics, are provided in Section~\ref{sec:setup}.



\subsection{Reviewer Assessor}
\label{sec:stage_1}

Given a target paper $p$ and a candidate reviewer $r$, the \methodAbbre{}-Assessor produces a binary label $\hat{y} \in \{0, 1\}$ indicating whether $r$ is suitable to review $p$.
Specifically, the model generates a structured chain-of-thought with four parts: 
(1)~\emph{required expertise}, a list of expertise dimensions 
identified from the target paper; (2)~\emph{evidence clues}, 
specific publications from the reviewer's history linked to each dimension; (3)~\emph{reasoning summary}, a synthesis of 
the evidence; and (4)~\emph{final label}, the suitability decision.
This structure grounds each prediction in explicit evidence and makes the decision process transparent. 
Since gold suitability labels are unavailable at scale, we train the model with reinforcement learning using rewards from an LLM judge guided by paper-specific expertise rubrics. 
We first describe the rubric construction procedure, then detail the reward design.

\paragraph{Rubric construction.}
Each paper-specific expertise rubric specifies the expertise dimensions required to review the target paper and their relative importance.
Formally, a rubric is defined as $u_p = \{(c_j, w_j)\}_{j=1}^{k}$, where each criterion 
$c_j = (t_j, d_j)$ consists of a title $t_j$ and a description $d_j$ specifying one expertise dimension, and $w_j$ denotes its importance weight.
Criteria are divided into two tiers: \emph{Core} criteria ($w_j = 5$) represent 
expertise essential for evaluating the paper's central contribution, while \emph{Secondary} criteria ($w_j \in \{3, 4\}$) capture supporting knowledge.
Each rubric contains $3$--$6$ criteria with 1--2 Core entries, as a small number of well-differentiated dimensions yields more reliable LLM-judge evaluation than a long list of 
overlapping items~\citep{RaR}.
Rubrics are generated by prompting an LLM with the paper's title, abstract, and introduction, along with human-written exemplars. 
Prompts and a rubric example are provided in Appendices~\ref{app:stage1-prompts} 
and~\ref{app:human-written-rubric}.

\paragraph{Reward design.}

The \methodAbbre{}-Assessor generates a structured chain-of-thought before producing a final label. Two failure modes can arise in this process: the analysis may fail to identify the expertise dimensions critical to the paper, or may assert relevance without citing specific publications as evidence; and the final label may contradict the analysis itself.
The reward signal is designed to penalize both.
We decompose it into two components---\emph{rubric coverage}, which measures how thoroughly the analysis addresses the rubric criteria with evidence from the reviewer's publications, and \emph{decision consistency}, which measures whether the final label is consistent with the evidence presented in the analysis---and combine them through a gating mechanism so that the model is rewarded only when its decision is supported by a thorough, evidence-grounded assessment. Both components are scored by the LLM judge in a single pass over the model output and rubric.


\textbf{Rubric coverage.}
For each rubric entry $(c_j, w_j) \in u_p$, the LLM judge assigns a binary indicator $m_j \in \{0, 1\}$, setting $m_j = 1$ if the output addresses the expertise dimension 
specified by $c_j$ with grounded evidence---either by citing relevant publications from the reviewer's history or by explicitly noting the absence of such evidence. 
The overall coverage score is the importance-weighted fraction of covered criteria:
\begin{equation}
  s_{\text{cov}} 
  = \frac{\sum_{j=1}^{k} w_j\, m_j}
         {\sum_{j=1}^{k} w_j}.
  \label{eq:coverage}
\end{equation}


\textbf{Decision consistency.}
A positive prediction ($\hat{y} = 1$) requires evidence that the reviewer covers at least one Core criterion and at least one additional criterion, ensuring that the decision reflects both central expertise and sufficient breadth; a negative prediction ($\hat{y} = 0$) is consistent when this condition is not met. The LLM judge assigns 
$s_{\text{dec}} \in \{0, 1\}$, with $s_{\text{dec}} = 1$ if 
$\hat{y}$ is consistent with this rule and $s_{\text{dec}} = 0$ otherwise.

\textbf{Gated reward.}
The final reward gates coverage on consistency:
\begin{equation}
  R = s_{\text{dec}} \cdot s_{\text{cov}}.
  \label{eq:reward}
\end{equation}
When $s_{\text{dec}} = 0$, the reward is zero regardless of coverage, ensuring that no credit is given to outputs whose final label contradicts their own analysis.
The full prompt templates for both the policy model and the LLM judge are provided in Appendix~\ref{app:stage1-prompts}.

\paragraph{RL training.}
We optimize the \methodAbbre{}-Assessor with GRPO~\citep{grpo}.
For each input prompt $x^i$, the LLM policy $\pi_{\theta}$ samples a group of $G$ outputs $\{o_j^i\}_{j=1}^{G}$, and each output receives a reward $r_j^i=R(o_j^i)$. We compute the group-wise advantage as:
\begin{equation}
    \hat A_j^i =
    \frac{
    r_j^i - \text{mean}\left(\{r_l^i\}_{l=1}^G\right)
    }{
    \text{std}\left(\{r_l^i\}_{l=1}^G\right)
    } .
\end{equation}
The same advantage $\hat A_j^i$ is applied to every token in output $o_j^i$. Following DAPO~\citep{dapo}, we adopt the clipped surrogate objective with the clip-higher strategy to ensure update stability while encouraging exploration. Specifically, for each token $o_{j,t}^i$, we define the policy ratio as:
\begin{equation}
\rho_{j,t}^i =
\frac{
\pi_{\theta}(o_{j,t}^i \mid x^i,o_{j,<t}^i)
}{
\pi_{\theta_{\mathrm{old}}}(o_{j,t}^i \mid x^i,o_{j,<t}^i)
}.
\end{equation}
The token-level objective is then defined as:
\begin{equation}
\begin{aligned}
l_{j,t}^i =
\min
\Big(
&\rho_{j,t}^i \hat A_j^i,
\operatorname{clip}
\big(
\rho_{j,t}^i,
1-\epsilon_{\mathrm{l}},
1+\epsilon_{\mathrm{h}}
\big)
\hat A_j^i
\Big).
\end{aligned}
\label{eq:grpo-token-objective}
\end{equation}
The final RL objective averages the token-level surrogate over generated tokens and adds a KL regularization term toward the reference policy:
\begin{equation}
\begin{aligned}
\mathcal{J}_{\mathrm{GRPO}}(\theta)
=&\;
\mathbb{E}_{\{o_j^i\}_{j=1}^{G}\sim \pi_{\theta_{\mathrm{old}}}}
\left[
\frac{
\sum_{j=1}^{G}\sum_{t=1}^{|o_j^i|} l_{j,t}^i
}{
\sum_{j=1}^{G}|o_j^i|
}
\right] \\
&\;-
\beta \mathcal{D}_{\mathrm{KL}}
(\pi_{\theta}\Vert\pi_{\mathrm{ref}}).
\end{aligned}
\label{eq:grpo-objective}
\end{equation}
Here, $\mathcal{D}_{\mathrm{KL}}$ is computed using the low-variance KL estimator (k3) between $\pi_\theta$ and $\pi_{\mathrm{ref}}$.

\subsection{\methodAbbre{}-Retriever}
\label{sec:stage_2}
Applying the \methodAbbre{}-Assessor to every candidate is prohibitively expensive at conference scale. 
We therefore distill its predictions into an embedding-based retriever. 
Using a separate subset of candidate pairs (constructed as in Section~\ref{sec:DataConstruction}), we annotate each pair with the trained assessor to obtain pseudo-labeled tuples $(p, r, \hat{y})$ , which are then converted into preference triplets for training.

\paragraph{Preference Data Construction}
Each training instance is a preference triplet 
$(a, c^+, c^-)$, where $a$ is an anchor, $c^+$ a suitable candidate ($\hat{y}=1$), and $c^-$ an unsuitable candidate ($\hat{y}=0$). 
For each anchor, every positive candidate is paired with every negative candidate. 
We construct triplets under two complementary views: (1) in the \emph{paper-centric} view, the anchor is a target paper and 
the candidates are reviewers; (2) in the \emph{reviewer-centric} view, the anchor is a reviewer and the candidates are papers. 
Both views share the same triplet format and are trained jointly for more robust representations~\citep{RATE, dual}.

\paragraph{Paper-conditioned reviewer profiles.}
\label{sec:reviewerprofiles}
Training on the preference triplets above requires encoding both papers and reviewers into a shared embedding space. Since the MERIT-Assessor evaluates whether a reviewer's publications collectively cover multiple expertise dimensions, the retriever's reviewer representation should capture the same cross-publication patterns. However, prior methods~\citep{CoF,CMU} score each publication against the target paper independently and aggregate the results (e.g., via percentile pooling), discarding inter-publication dependencies.
We instead construct a paper-conditioned reviewer profile by selecting the $K$ publications from $H_r$ most relevant to $p$, ranked by cosine similarity under the backbone embedding model $f_0$ (before fine-tuning):
\begin{equation}
  s_p(h_m) = \cos\bigl(f_0(p),\, f_0(h_m)\bigr).
  \label{eq:profile_sim}
\end{equation}
Restricting to the top $K$ publications keeps the input compact while filtering out unrelated work.
The selected publications are concatenated with a task-specific instruction prefix (e.g., ``For the Author (Query): Represent this author's publication history and expertise for finding relevant academic papers.'') and encoded as a single sequence to produce the reviewer representation.
We use the frozen $f_0$ for publication selection throughout both training and evaluation, ensuring consistency independent of 
the retriever's learned parameters.

\begin{table*}[t]
\centering
\small
\caption{
Reviewer suitability classification results on LR-Bench.
Best and second-best results are marked in bold and underlined, respectively.
}
\label{tab:suitability-classification}
\resizebox{0.75\textwidth}{!}{
\begin{tabular}{lccccc} 
\toprule
& \multicolumn{5}{c}{\textbf{LR-Bench}} \\
\cmidrule(lr){2-6}
\textbf{Model} 
& \textbf{Acc.} ($\uparrow$)
& \textbf{B.Acc.} ($\uparrow$)
& \textbf{P.} ($\uparrow$)
& \textbf{R.} ($\uparrow$)
& \textbf{F1} ($\uparrow$) \\

\midrule
\rowcolor[HTML]{F2F2F2}
\multicolumn{6}{l}{\textbf{Direct prompting}} \\

Qwen3.5-Plus 
& 59.18\% & 64.41\% & 50.52\% & \textbf{95.67\%} & 66.12\% \\
DeepSeek-V3.2 
& 63.11\% & 66.93\% & 53.40\% & \underline{89.76\%} & 66.96\% \\
\midrule
\rowcolor[HTML]{F2F2F2}
\multicolumn{6}{l}{\textbf{Expertise-aware prompting}} \\
Qwen3-4B 
& 67.08\% & 68.81\% & 57.61\% & 78.45\% & 66.15\% \\
Qwen3.5-Plus 
& 67.87\% & 70.39\% & 57.71\% & 85.43\% & \underline{68.89\%} \\
DeepSeek-V3.2 
& \underline{69.51\%} & \underline{71.00\%} & \underline{60.06\%} & 79.92\% & 68.58\% \\
\midrule
\methodAbbre{}-Assessor-4B 
& \textbf{71.64\%} & \textbf{73.11\%} & \textbf{62.09\%} & 81.89\% & \textbf{70.63\%} \\
\bottomrule
\end{tabular}
}
\label{tab:results_0}
\end{table*}

\paragraph{Dual-view Preference Alignment}

We optimize an embedding model $f_\theta$, initialized from $f_0$, with a multi-objective loss over the preference triplets from both views.
The matching score between an anchor $a$ and a 
candidate $c$ is defined as:
\begin{equation}
  s(a, c) = \cos\bigl(f_\theta(a),\, f_\theta(c)\bigr).
  \label{eq:matching}
\end{equation}
The pairwise ranking loss encourages the model to score the suitable candidate higher than the unsuitable one:
\begin{equation}
  \mathcal{L}_{\text{pair}} 
  = -\log \sigma\!\left(
      \frac{s(a, c^+) - s(a, c^-)}{\tau}
    \right),
  \label{eq:pair}
\end{equation}
where $\tau$ is a temperature hyperparameter. The in-batch contrastive loss provides additional negative signal by contrasting the positive candidate against all candidates in the mini-batch $\mathcal{B}$:
\begin{equation}
  \mathcal{L}_{\text{nce}} 
  = -\log 
    \frac{\exp\bigl(s(a, c^+)/\tau\bigr)}
         {\sum_{c \in \mathcal{B}} 
          \exp\bigl(s(a, c)/\tau\bigr)}.
  \label{eq:nce}
\end{equation}
The final objective combines both terms:
\begin{equation}
\mathcal{L}
=
\mathcal{L}_{\mathrm{pair}}
+
\lambda_{\mathrm{nce}}\mathcal{L}_{\mathrm{nce}}.
\label{eq:retriever-final-loss}
\end{equation}
We fine-tune $f_\theta$ with 
LoRA~\citep{lora} for parameter-efficient adaptation while preserving the pretrained representations. 
Training hyperparameters are provided in 
Appendix~\ref{app:training-hyperparameters}.

\section{Experiments}
\label{sec:experiments}
Our experiments address three questions. 
\textbf{(Q1)} Does rubric-guided reward training improve suitability classification over general-purpose LLM prompting? (Section~\ref{sec:result1}) 
\textbf{(Q2)} Does distilling the \methodAbbre{}-Assessor's predictions into an embedding-based retriever improve reviewer retrieval? (Section~\ref{sec:result2}) 
\textbf{(Q3)} How does the quality of the pseudo-label source affect downstream retrieval? (Section~\ref{sec:Ablation-study}) 
We additionally report sensitivity to reviewer profile size in Appendix~\ref{sec:profile-size}, cost analysis in Appendix~\ref{sec:cost-analysis} and human validation of the generated rubrics and LLM judge in Appendix~\ref{sec:human-validation}.

\subsection{Experimental Setup}
\label{sec:setup}
\paragraph{Training data.}
All training data are drawn from the RATE corpus~\citep{RATE}, with papers appearing in any evaluation benchmark removed to prevent data leakage. We use two disjoint splits for the two stages of our framework, constructed by a shared procedure: we recover each paper's introduction from ar5iv~\citep{ar5iv} and discard papers for which no 
introduction is available; we restrict the reviewer pool to authors with at least three publications and exclude authors whose publication history contains the target paper itself; and for each target paper, we rank all eligible reviewers with the RATE retrieval model and retain the top 5 as candidates.

For training the \methodAbbre{}-Assessor 
(Section~\ref{sec:stage_1}), we randomly sample 2,000 papers, of which 1,497 have available introductions. We generate one expertise rubric per paper using 
Qwen3-Max-Thinking, yielding 7,485 rubric-equipped training tuples $(p, r, u_p)$.
For training the \methodAbbre{}-Retriever (Section~\ref{sec:stage_2}), we sample 
a separate set of 6,000 papers (5,374 with available 
introductions) and construct 26,870 candidate paper--reviewer 
pairs using the same procedure. These pairs are annotated by 
the trained suitability model and converted into dual-view 
preference triplets following Section~\ref{sec:stage_2}, yielding 8,720 
paper-centric and 7,208 reviewer-centric triplets.

\begin{table*}[t]
\caption{
Reviewer retrieval results on LR-Bench and CMU Gold.
We report expertise-aligned loss and pairwise accuracy. Best and second-best results are marked in bold and underlined, respectively.
}
  \centering
  \resizebox{1.0\textwidth}{!}{
    \begin{tabular}{l c c c g c c c  g}
      \toprule
      
      & \multicolumn{4}{c}{\textbf{Loss} ($\downarrow$)} & \multicolumn{4}{c}{\textbf{Acc.} ($\uparrow$)} \\
      
      \cmidrule(lr){2-5} \cmidrule(lr){6-9}
      
      \textbf{Algorithm} & \textbf{LR-PC} & \textbf{LR-RC} & \textbf{Gold} & \textbf{Avg.} & \textbf{LR-PC} & \textbf{LR-RC} & \textbf{Gold} & \textbf{Avg.} \\
      \midrule
      \rowcolor[HTML]{F2F2F2}
      \multicolumn{9}{l}{\textbf{Statistical-based Methods}} \\
      TPMS &  0.2920&  0.2322&  0.2811&  0.2684& 66.22\% & 72.01\% & 71.89\% & 70.04\% \\
      \rowcolor[HTML]{F2F2F2}
      \multicolumn{9}{l}{\textbf{LLM-based Methods}} \\
      DeepSeek-V3.2 &  0.2736&  0.2348& 0.2237 &  0.2440& 50.34\% & 54.04\% & 77.36\% & 60.58\% \\
      Qwen3-Max     &  0.2698&  0.2289& 0.2246&  0.2411& 47.65\% & 55.01\% & 77.54\% & 60.07\% \\
      \midrule
      \rowcolor[HTML]{F2F2F2}
      \multicolumn{9}{l}{\textbf{Embedding-based Methods}} \\
      BERTScore & 0.2691 & 0.3380 & 0.3414 &  0.3162& 66.22\% & 62.36\% & 65.86\% & 64.81\% \\
      CoF & 0.2996 & 0.2175 & 0.2564 &  0.2578& 64.19\% & 73.34\% & 74.36\% & 70.63\% \\
      SPECTER   & 0.2118 & 0.2396 & 0.2851 &  0.2455& \underline{72.97}\% & 71.29\% & 71.49\% & 71.92\% \\
      SPECTER2 PRX & 0.1966 & 0.2175 & 0.2436 &  0.2192& \underline{72.97}\% & 72.26\% & 75.64\% & 73.62\% \\
      SciNCL    & 0.2042 & 0.1938 & 0.2663 &  0.2214& 72.30\% & 75.03\% &  73.37\% & 73.57\% \\
      RATE-8B&  \underline{0.1851}&  \underline{0.1857}&  \underline{0.1991}&  \underline{0.1900}& \textbf{75.00}\% & \underline{75.63}\% &\underline{80.09}\% &\underline{76.91}\% \\
      \midrule
       \methodAbbre{}-Retriever-8B  &   \textbf{0.1698}&  \textbf{0.1783}&  \textbf{0.1842}&  \textbf{0.1774}&  \textbf{75.00}\%&\textbf{76.84}\%& \textbf{81.58}\% & \textbf{77.81}\% \\
      \bottomrule
    \end{tabular}
  }
  \label{tab:main_results}
\end{table*}

\paragraph{Training details.}
For the \methodAbbre{}-Assessor, we initialize the policy model from Qwen3-4B~\citep{qwen3} and optimize with GRPO using the verl framework~\citep{verl}, with DeepSeek-V3.2-Thinking~\citep{deepseekv32} as the LLM judge.
For the \methodAbbre{}-Retriever, we initialize from Qwen3-Embedding-8B~\citep{qwen3embedding} and fine-tune with LoRA~\citep{lora}. 
The same embedding backbone is used to construct paper-conditioned reviewer profiles during both training and evaluation, following Section~\ref{sec:reviewerprofiles}. 
Detailed experimental settings are provided in Appendix~\ref{app:DetailedExperimentalSettings}.

\paragraph{Evaluation benchmarks.}
For suitability classification, we use LR-Bench~\citep{RATE}, where each reviewer–paper pair carries a self-assessed expertise rating on a 1–5 scale. We binarize these ratings by treating scores of 4 and 5 as positive and the remaining scores as negative. 
To ensure consistency with the training input, we also recover each target paper's introduction and discard examples where it is unavailable, yielding 810 labeled pairs split into 200 for validation and 610 for testing.

For reviewer retrieval, we evaluate on LR-Bench and the CMU gold standard dataset~\citep{CMU}. Both datasets provide sparse expertise ratings based on human annotation; we derive pairwise preferences from these ratings following their original evaluation protocols. LR-Bench yields both paper-centric (PC) and reviewer-centric (RC) pairs, while CMU yields only reviewer-centric pairs. 
We reserve 30\% of LR-Bench preference pairs for validation and use the remaining 70\% for testing. CMU is used entirely for testing.

\subsection{Metrics}
For suitability classification, we report accuracy (Acc.), balanced accuracy (B.Acc.), precision (P.), recall (R.), and F1 score.

For reviewer retrieval, following \citet{CMU} and \citet{RATE}, we adopt a normalized ranking loss $\mathcal{L} \in [0,1]$ as our primary metric.
Given a set of preference pairs $\mathcal{P}$, where each pair $(x, y)$ shares the same anchor
and satisfies $\epsilon_x > \epsilon_y$ under ground-truth expertise labels, the loss penalizes misordered pairs in proportion to the gap between the two expertise ratings:
\begin{equation}
\mathcal{L}
=
\frac{
  \sum_{(x,y) \in \mathcal{P}}
  \mathcal{I}(s_x < s_y) \cdot |\epsilon_x - \epsilon_y|
}{
  \sum_{(x,y) \in \mathcal{P}}
  |\epsilon_x - \epsilon_y|
},
\label{eq:expertise-aligned-loss}
\end{equation}
where $\mathcal{I}$ is the indicator function, and $s$ denotes the model-predicted score, and the loss is normalized to $[0,1]$ by the sum of all pairwise expertise gaps.
In addition, we report pairwise accuracy, the fraction of pairs where the model correctly orders the two candidates by ground-truth rating.

\subsection{Baselines}

\paragraph{Suitability classification.}
We consider two categories of LLM-based baselines.
\textbf{(1) Direct prompting} predicts reviewer suitability from the target paper and reviewer profile without explicit expertise decomposition.
\textbf{(2) Expertise-aware prompting}  follows the same structured assessment process as our policy model---expertise identification, evidence matching, and final prediction---but without RL training.
We evaluate direct prompting with Qwen3.5-Plus and
DeepSeek-V3.2, and expertise-aware prompting with Qwen3-4B, Qwen3.5-Plus, and DeepSeek-V3.2.
All baselines use greedy decoding with thinking mode enabled.
\paragraph{Reviewer retrieval.}
We compare against three categories of baselines.
\textbf{(1) Statistical method.}
We include TPMS~\citep{tpms}, a standard reviewer-assignment baseline based on TF-IDF similarity between the target paper and the reviewer's publications.
\textbf{(2) Embedding-based models.}
We compare with BERTScore~\citep{bertscore}, a common text similarity metric based on contextual token embeddings.
We also include SPECTER~\citep{specter},
SciNCL~\citep{scincl}, and SPECTER2 with the PRX
adapter~\citep{SciRepEval}, all of which are scientific
document encoders trained on citation links between papers.
In addition, we compare with CoF~\citep{CoF}, a factor-aware reviewer-assignment framework that combines semantic, topical, and citation signals; and RATE-8B~\citep{RATE}, a weakly supervised reviewer retriever that combines high-confidence proxy signals with keyword-based reviewer profiling.
\textbf{(3) LLM-based methods.}
We prompt DeepSeek-V3.2 and Qwen3-Max to score paper--reviewer affinity in a zero-shot setting, and use the resulting scores for pairwise ranking.
Additional implementation details and LLM-baseline prompts are provided in Appendices~\ref{app:baseline-details} and~\ref{app:baseline-prompts}.

\subsection{Main Results}
\subsubsection{Reviewer suitability classification (Q1)}
\label{sec:result1}
As shown in Table~\ref{tab:suitability-classification}, the \methodAbbre{}-Assessor achieves the best accuracy, balanced accuracy, precision, and F1 score, despite using a smaller 4B backbone than several larger prompting baselines. 
The most direct comparison is with Qwen3-4B under the same expertise-aware format but without training: our model improves Acc. from 67.08\% to 71.64\% and F1 from 66.15\% to 70.63\%, confirming that rubric-guided reinforcement learning teaches the model to make more discriminative judgments beyond what the structured format alone provides.

The prompting results also reveal a systematic positive prediction bias in direct LLM judgments. Without structured expertise decomposition, models tend to predict nearly all candidates as suitable — Qwen3.5-Plus reaches 95.67\% recall but only 50.52\% precision. Expertise-aware prompting mitigates this bias by requiring the model to identify expertise dimensions and match them against the reviewer's publications before deciding. This decomposition alone, without any training, lifts Qwen3-4B above larger direct-prompting models in both accuracy and precision, indicating that the structured assessment format provides a useful inductive bias.
These results establish the trained assessor as a stronger pseudo-label source; we next examine whether this improvement translates into better downstream retrieval.

\subsubsection{Reviewer retrieval (Q2)}
\label{sec:result2}
Table~\ref{tab:main_results} shows that the \methodAbbre{}-Retriever achieves the best or tied-best performance on every metric, reducing average expertise-aligned loss from 0.1900 to 0.1774 and raising average pairwise accuracy from 76.91\% to 77.81\% compared with RATE-8B, the strongest baseline.
Since both models share the same embedding backbone and retriever architecture, 
this improvement suggests that higher-fidelity criterion-level suitability supervision provides more effective training signals than RATE's proxy-derived weak supervision.

Among embedding-based baselines, citation-informed encoders (SPECTER, SPECTER2-PRX, SciNCL) consistently outperform BERTScore, confirming that citation-aware pretraining provides a stronger foundation for this task. 
RATE-8B further improves over these encoders through dedicated training with high-confidence proxy-derived weak supervision and keyword-based reviewer profiling, but its reliance on proxy signals limits the fidelity of its supervision. 
Notably, TPMS remains competitive despite relying solely on TF-IDF similarity without any learned representations, achieving 70.04\% average accuracy---higher than both zero-shot LLM methods and close to CoF, which integrates semantic, topical, and citation signals. 

Zero-shot LLM scoring exhibits a distinctive pattern: DeepSeek-V3.2 and Qwen3-Max achieve relatively low expertise-aligned loss (0.2440 and 0.2411) but low pairwise accuracy (60.58\% and 60.07\%). This gap indicates that while LLMs can identify clearly unsuitable reviewers, they struggle with fine-grained pairwise ordering among topically related candidates---precisely the regime where criterion-level matching matters most. Their accuracy also varies sharply across benchmarks, exceeding 77\% on CMU Gold but falling below 55\% on LR-Bench, suggesting sensitivity to the distribution of candidate difficulty.

\begin{table*}[!t]
\centering
\footnotesize
\setlength{\tabcolsep}{4pt}
\caption{Effect of the Stage-1 pseudo-label source on retrieval
performance (LR-Bench). All variants annotate the same candidate
paper--reviewer pairs and use the same retriever training procedure. Avg.\ denotes the macro-average over PC and RC.}
\label{tab:lr_stage1_model_ablation}
\resizebox{0.94\textwidth}{!}{
\begin{tabular}{lccccccc}
\toprule
\multirow{2}{*}{\textbf{Pseudo-label Source}} 
& \multirow{2}{*}{\textbf{\#Triplets (PC, RC)}} 
& \multicolumn{3}{c}{\textbf{Loss} ($\downarrow$)} 
& \multicolumn{3}{c}{\textbf{Acc.} ($\uparrow$)} \\
\cmidrule(lr){3-5} \cmidrule(lr){6-8}
& & \textbf{PC} & \textbf{RC} & \textbf{Avg.} 
& \textbf{PC} & \textbf{RC} & \textbf{Avg.} \\
\midrule
Qwen3-4B 
& (12077, 9259) 
& 0.2004 & 0.1879 & 0.1942 
& 72.30\% & 74.79\% & 73.55\% \\

DeepSeek-V3.2 
& (7037, 6112) 
& \underline{0.1813} & \underline{0.1805} & \underline{0.1809} 
& \underline{74.32\%} & \underline{75.87\%} & \underline{75.10\%} \\

\methodAbbre{}-Assessor-4B 
& (8720, 7208) 
& \textbf{0.1698} & \textbf{0.1783} & \textbf{0.1741} 
& \textbf{75.00\%} & \textbf{76.84\%} & \textbf{75.92\%} \\
\bottomrule
\end{tabular}
}
\end{table*}




\subsection{Effect of Pseudo-label Quality (Q3)}
\label{sec:Ablation-study}

Since Stage-2 retrieval relies on pseudo-labels generated by the Stage-1 annotator, we study how pseudo-label quality affects downstream retrieval performance.
To examine this, we vary the Stage-1 annotator while keeping the retrieval training pipeline and candidate pairs unchanged.
We compare three pseudo-label sources: (1) the base Qwen3-4B model with expertise-aware prompting but without RL training, (2) DeepSeek-V3.2, the strongest general-purpose prompting baseline (Table~\ref{tab:results_0}), and (3) the trained
\methodAbbre{}-Assessor. All variants are evaluated on LR-Bench.

As shown in Table~\ref{tab:lr_stage1_model_ablation}, downstream retrieval performance improves monotonically as the annotator becomes stronger. Although different annotators produce different label distributions and therefore different numbers of preference triplets, Qwen3-4B generates the most triplets yet yields the weakest retriever, suggesting that retrieval performance is more closely associated with pseudo-label quality than quantity. Overall, the consistent trend indicates that higher-fidelity suitability predictions provide more effective supervision for the retriever.

\section{Conclusion}

We identify a key challenge in automatic reviewer assignment: scalable supervision signals lack fidelity to true reviewer suitability, while high-fidelity signals are difficult to scale. To address this, we proposed \methodAbbre{}, a two-stage framework that trains a reviewer assessor with paper-specific expertise rubrics and distills its predictions into an efficient embedding-based retriever for large-scale assignment. Experiments showed that the trained 4B assessor produces more discriminative suitability judgments than larger general-purpose LLMs, and that distilling its predictions yields consistent retrieval gains across two benchmarks.
These results demonstrate that rubric-guided, criterion-level supervision can better balance fidelity and scalability in automatic reviewer assignment.

\section{Limitations}
\label{sec:Limitations}

Despite the promising results, our work has several limitations.
First, the construction of the training set depends on RATE's recall for candidate mining, which may bias the learned suitability boundaries when RATE fails to retrieve relevant candidates.

Second, our reviewer profiles treat all publications equally without modeling author-order signals such as first- or last-author versus middle-author contributions. In fields where author position is a proxy for contribution depth, this may lead the model to overestimate reviewer expertise based on publications with peripheral involvement.

Third, our evaluation focuses on expertise matching in isolation and does not account for practical constraints in real-world reviewer assignment, such as reviewer availability, workload balance, conflicts of interest, and expertise scarcity.

Fourth, all experiments are conducted on computer science venues, and the rubric structure---a small number of weighted, tiered
criteria---reflects the expertise patterns typical of this domain. Whether this structure generalizes to fields with different reviewing norms remains to be validated.
\section*{Ethical Consideration}

\paragraph{Potential Risks.}
\methodAbbre{} is designed to assist reviewer assignment in academic
peer review. Although the system aims to improve matching quality, we
recognize several potential risks. First, the suitability model and
rubric generator may inherit biases from their underlying LLMs,
potentially favoring certain research paradigms or methodologies over others. Second, we emphasize that \methodAbbre{}
provides ranked suggestions, not final decisions, and is intended to
assist rather than replace human judgment in reviewer assignment.
Third, aggregating a researcher's publication history into a structured
expertise profile constitutes an automated capability assessment. All publications used are publicly available and drawn from
existing research datasets released for academic use, and no private
or personally identifiable information beyond public authorship
metadata is involved.

\paragraph{Licenses, Intended Use, and Sensitive Information.}
All data used in this study are derived from publicly available academic resources. Our training data are drawn from the RATE
corpus, which is constructed from public
publications on arXiv. The evaluation benchmarks,
LR-Bench and CMU Gold, are publicly released for research purposes. 
These datasets contain no
sensitive personal information beyond publicly available publication metadata.
For model training, we initialize from
Qwen3-4B and
Qwen3-Embedding-8B, both released under
open-source licenses. We use the verl
framework (Apache 2.0 License) for GRPO
training. For repositories where a specific license was not explicitly
provided, we have used them strictly in accordance with their intended
research purposes. Our code will be released under an open-source
license to facilitate reproducibility.

\bibliography{custom}
\clearpage
\appendix
\section*{Appendix}

\section{Additional Experiments}

\subsection{Sensitivity to profile size.}
\label{sec:profile-size}

We analyze sensitivity to $K$, the number of reviewer publications used to construct paper-conditioned profiles, holding all other settings fixed and applying the same $K$ during both training and evaluation.
As shown in Table~\ref{tab:lr_topk_sensitivity}, performance follows an inverted-U pattern: accuracy rises from $K{=}1$ (72.93\%) to a peak at $K{=}3$ (75.92\%, 0.1741 loss), then declines at $K{=}4$ and $K{=}5$ (73.04\%).
Too few publications leave reviewer expertise underspecified,
while too many introduce marginally relevant work that dilutes the profile signal. 
We adopt $K{=}3$ as the default setting.
\begin{table}[H]
\centering
\scriptsize
\setlength{\tabcolsep}{4pt}
\caption{Sensitivity to the number of selected reviewer publications
$K$. We report expertise-aligned loss and pairwise accuracy on
LR-Bench.}
\resizebox{\columnwidth}{!}{
\begin{tabular}{ccccccc}
\toprule
\multirow{2}{*}{$K$} & \multicolumn{3}{c}{Loss ($\downarrow$)} & \multicolumn{3}{c}{Acc. ($\uparrow$)} \\
\cmidrule(lr){2-4} \cmidrule(lr){5-7}
& PC & RC & Avg. & PC & RC & Avg. \\
\midrule
1 & 0.2118  & 0.1930  & 0.2024   & 70.95\% & 74.91\% & 72.93\% \\
2 & 0.2195  & \underline{0.1805}  & 0.2000   & 70.95\% & \underline{76.56}\% & 73.76\% \\
3 & \textbf{0.1698}  & \textbf{0.1783}  & \textbf{0.1741}   & \textbf{75.00\%} & \textbf{76.84\%} & \textbf{75.92\%} \\
4 & \underline{0.1813}  & 0.1953  & \underline{0.1883}   & \textbf{75.00}\% & 75.51\% & \underline{75.26}\% \\
5 & 0.2004  & 0.2078  & 0.2041   & \underline{72.97}\% & 73.10\% & 73.04\% \\
\bottomrule
\end{tabular}
}
\label{tab:lr_topk_sensitivity}
\end{table}

\subsection{Cost Analysis}
\label{sec:cost-analysis}

To assess the practical efficiency of \methodAbbre{}, we analyze its cost along two axes: one-time offline training and deployment.

\begin{table}[H]
\caption{End-to-end offline cost of the \methodAbbre{} pipeline.}
\centering
\small
\begin{tabular}{lcc}
\toprule
\textbf{Stage} & \textbf{Compute} & \textbf{API cost} \\
\midrule
RATE candidate mining       & $\approx$21 GPU-h  & --       \\
Rubric generation           & --                  & \$23.58  \\
Rubric-guided judging       & --                  & \$78.57  \\
GRPO training               & $\approx$245 GPU-h & --       \\
Assessor pseudo-labeling    & $\approx$114 GPU-h  & --       \\
Retriever distillation      & $\approx$6 GPU-h    & --       \\
\midrule
\textbf{Total}              & $\approx$\textbf{386 GPU-h} & \textbf{\$102.15} \\
\bottomrule
\end{tabular}
\label{tab:offline-cost}
\end{table}

\paragraph{Offline training cost.}
Table~\ref{tab:offline-cost} breaks down the full offline pipeline of \methodAbbre{}.
GRPO training dominates the compute budget, while
rubric-guided judging accounts for most of the API cost. However, the entire cost is incurred once: after the retriever is distilled, no further LLM calls or policy-model inference is required at deployment.

\begin{table}[H]
\caption{Deployment cost on 1{,}000 randomly sampled preference triplets using a single NVIDIA A800-80GB GPU.}
\centering
\small
\setlength{\tabcolsep}{3.6pt}
\begin{tabular}{lccccc}
\toprule
& \textbf{Prep.} & \textbf{Infer.} & \textbf{LLM} & \textbf{GPU} & \textbf{Total} \\
& (s) & (s) & cost & cost & cost \\
\midrule
RATE  & 679 & 220 & \$1.57 & \$0.06 & \$1.63 \\
\methodAbbre{} & 60 & 713 & \$0.00 & \$0.20 & \$0.20 \\
\bottomrule
\end{tabular}
\label{tab:inference-cost}
\end{table}

\paragraph{Deployment cost.}
At deployment, \methodAbbre{} runs only the embedding retriever, whereas RATE-8B additionally requires generative-LLM-based keyword extraction for reviewer profiles.
We benchmark both methods on the same 1{,}000 randomly sampled preference triplets using one NVIDIA A800-80GB GPU.\footnote{GPU rental cost is estimated
at \$0.93 per GPU-hour based on prevailing cloud rates at the time of writing.}
As shown in Table~\ref{tab:inference-cost}, RATE's keyword generation step dominates its wall time and cost, with the generative-LLM API fee accounting for over 96\% of the total. \methodAbbre{} eliminates this dependency, reducing the deployment cost to GPU rental alone---an $8\times$ reduction overall.

\subsection{Human Validation}
\label{sec:human-validation}
Since \methodAbbre{} relies on automatically generated expertise
rubrics and an LLM judge to provide training supervision, we conduct two human validation studies to assess the reliability of these components.

\paragraph{Annotation protocol.}
The evaluation team consisted of five master's and doctoral students in computer science with relevant research experience. 
All annotators were compensated at a rate exceeding the local minimum wage and gave informed consent before participation. Annotators received task instructions and reference examples demonstrating the application of the evaluation criteria. 
All judgments were made independently, and human labels were determined by majority vote (at least 3 of 5 annotators).

\paragraph{Rubric quality.}
To validate the generated expertise rubrics, we randomly sampled
100 papers from the Stage-1 training set, yielding 405 rubric items (189 Core and 216 Secondary).
For each rubric item, annotators applied the evaluation criterion corresponding to its rubric tier:

\begin{itemize}[leftmargin=*, nosep]
    \item \textbf{Core}: whether it captures expertise essential for evaluating the paper's central contribution.
    \item \textbf{Secondary}: whether it captures supporting expertise relevant to evaluating the paper.
\end{itemize}

As shown in Table~\ref{tab:rubric-validation}, 89.6\% of rubric items were validated by majority vote, with Core and Secondary items reaching 91.0\% and 88.4\%, respectively.
The high validation rates across both tiers suggest that the generated criteria generally align with their intended roles.

\begin{table}[H]
\caption{Human validation of generated expertise rubrics. Validation labels are determined by majority vote among five annotators.}
\centering
\small
\resizebox{0.95\columnwidth}{!}{%
\begin{tabular}{lccc}
\toprule
\textbf{Type} & \textbf{Items} & \textbf{Validated} & \textbf{Not validated} \\
\midrule
Core      & 189 & 172 (91.0\%) & 17 (9.0\%)  \\
Secondary & 216 & 191 (88.4\%) & 25 (11.6\%) \\
\midrule
Overall   & 405 & 363 (89.6\%) & 42 (10.4\%) \\
\bottomrule
\end{tabular}%
}
\label{tab:rubric-validation}
\end{table}

\paragraph{Judge--human alignment.}
To assess the reliability of the LLM judge's evaluations, we sampled 100 paper--reviewer pairs from the training set and used the base Qwen3-4B with the training policy prompt to generate suitability assessments. The LLM judge and five annotators then independently evaluated these assessments for rubric coverage and decision consistency. We measured judge--human agreement using the human majority-vote labels as reference.

As shown in Table~\ref{tab:judge-alignment}, the LLM judge achieves over 82\% F1 on both rubric coverage and decision consistency. These results suggest that the judge provides a reasonably reliable supervision signal for training the \methodAbbre{}-Assessor.

\begin{table}[H]
\caption{LLM judge alignment with human evaluators on rubric coverage and decision consistency. Human reference labels are determined by majority vote among five annotators.}
\centering
\small
\resizebox{0.97\columnwidth}{!}{%
\begin{tabular}{lcccc}
\toprule
\textbf{Type} & \textbf{Acc.} & \textbf{P.} & \textbf{R.} & \textbf{F1} \\
\midrule
Core coverage      & 82.7\% & 82.0\% & 87.2\% & 84.5\% \\
Secondary coverage & 79.5\% & 79.0\% & 83.8\% & 81.3\% \\
Overall coverage   & 80.9\% & 80.4\% & 85.3\% & 82.8\% \\
\midrule
Decision consistency & 79.0\% & 76.1\% & 91.1\% & 82.9\% \\
\bottomrule
\end{tabular}%
}
\label{tab:judge-alignment}
\end{table}

\section{Detailed Experimental Settings}
\label{app:DetailedExperimentalSettings}

\subsection{Computing Facilities}
We train the \methodAbbre{}-Assessor with GRPO on
4$\times$ NVIDIA A800-80GB GPUs and fine-tune the \methodAbbre{}-Retriever on
2$\times$ NVIDIA A800-80GB GPUs. All evaluation experiments are
conducted on a single NVIDIA A800-80GB GPU.

\subsection{Training Hyperparameters}
\label{app:training-hyperparameters}

Tables~\ref{tab:stage1-hyperparameters}
and~\ref{tab:stage2-hyperparameters} list the main hyperparameters for
the \methodAbbre{}-Assessor and the \methodAbbre{}-Retriever, respectively.

\subsection{Baseline details}
\label{app:baseline-details}
Table~\ref{tab:baseline_resources} summarizes the implementation sources
and model checkpoints used for retrieval baselines. All baselines are
evaluated using their released checkpoints.

For baselines that score a reviewer through profile publications, we follow their original evaluation protocols whenever available, including the profile construction and score aggregation strategies, to ensure a fair comparison.
BERTScore, SPECTER, SPECTER2-PRX, and SciNCL are paper-level encoders
that produce similarities between individual document pairs. To obtain a
paper--reviewer score, we compute similarities between the target paper
and each of the reviewer's publications and aggregate via 75th-percentile pooling~\citep{75profile}.
For CoF and RATE, which
include their own reviewer-level scoring procedures---top-$3$ averaging
and keyword-based profile construction, respectively---we follow their
original protocols.

\begin{table}[htbp]
\centering
\caption{Main hyperparameters for training the  \methodAbbre{}-Assessor.}
\label{tab:stage1-hyperparameters}
\resizebox{0.5\textwidth}{!}{
\begin{tabular}{lcc}
\toprule
Parameter & Training & Inference \\
\midrule
clip\_low               & 0.2 & N/A \\
clip\_high              & 0.28 & N/A \\
learning\_rate          & $5e-7$ & N/A \\
max\_prompt\_length   & 5120 & N/A \\
max\_response\_length   & 4096 & 4096 \\
total\_epochs           &1 &N/A \\
train\_prompt\_batchsize & 16 & N/A \\
group\_size             & 8 & N/A \\
train\_prompt\_mini\_batchsize & 8 & N/A \\
temperature             & 1.0 & 0.0 \\
top\_p                  & 1.0 & N/A \\
\bottomrule
\end{tabular}
}
\end{table}

\begin{table}[htbp]
\centering
\small
\caption{Main hyperparameters for fine-tuning the \methodAbbre{}-Retriever.}
\label{tab:stage2-hyperparameters}
\resizebox{0.5\textwidth}{!}{
\begin{tabular}{lc}
\toprule
Parameter & Value \\
\midrule
Mixed precision & bf16 \\
Attention implementation & FlashAttention-2 \\
Batch size & 16 \\
Evaluation batch size & 4 \\
Gradient accumulation steps & 2 \\
Epochs & 1 \\
Learning rate & $4e-5$ \\
Warmup ratio & 0.01 \\
Temperature & 0.04 \\
$\lambda_{nce}$ & 0.5 \\
Max query length & 2048 \\
Max document length & 2048 \\
Patience & 100 \\
Evaluation frequency & 10 steps \\
Seed & 42 \\
LoRA rank & 32 \\
LoRA alpha & 64 \\
LoRA dropout & 0.06 \\
\bottomrule
\end{tabular}
}
\end{table}

\section{Large Language Model Prompts}
\label{app:prompts}

We provide the prompt templates used in our method and LLM-based
baselines. All prompts use placeholder variables (e.g.,
\texttt{\{\{paper\_title\}\}}) that are filled at inference time.

\subsection{Prompts for \methodAbbre{}-Assessor Training}
\label{app:stage1-prompts}

Stage~1 involves three prompts that operate in sequence: (1)~a rubric
generation prompt that produces paper-specific expertise criteria from
the target paper's title, abstract, and introduction
(Figure~\ref{fig:rubric-generation-prompt}); (2)~a policy-model prompt that elicits a structured chain-of-thought assessment of reviewer
suitability (Figure~\ref{fig:policy-prompt}); and (3)~an LLM-judge
prompt that evaluates each assessment against the generated rubric to
compute the gated reward (Figure~\ref{fig:judge-prompt}).

\begin{table}[ht]
\centering
\small
\caption{Implementation sources and model checkpoints for retrieval baselines.}
\label{tab:baseline_resources}
\resizebox{0.46\textwidth}{!}{
\begin{tabularx}{\columnwidth}{lX}
\toprule
\textbf{Method} & \textbf{Source / Checkpoint} \\
\midrule
TPMS 
& \href{https://github.com/niharshah/goldstandard-reviewer-paper-match}
{\texttt{niharshah/reviewer-paper-match}} \\

BERTScore 
& \href{https://github.com/Tiiiger/bert_score}
{\texttt{Tiiiger/bert\_score}} \\

SPECTER 
& \href{https://huggingface.co/allenai/specter}
{\texttt{allenai/specter}} \\

SciNCL 
& \href{https://huggingface.co/malteos/scincl}
{\texttt{malteos/scincl}} \\

SPECTER2-PRX 
& \href{https://huggingface.co/allenai/specter2}
{\texttt{allenai/specter2}} \\

CoF 
& \href{https://github.com/yuzhimanhua/CoF}
{\texttt{yuzhimanhua/CoF}} \\

RATE-8B 
& \href{https://github.com/Gnociew/RATE-Reviewer-Assignment}
{\texttt{Gnociew/RATE-Reviewer-Assignment}} \\
\bottomrule
\end{tabularx}
}
\end{table}

\subsection{Prompts for LLM-Based Baselines}
\label{app:baseline-prompts}

We use baseline prompts corresponding to the two evaluation settings.
For reviewer suitability classification, we evaluate two prompting
strategies: direct prompting predicts a binary label without explicit
expertise decomposition (Figure~\ref{fig:direct-prompting-prompt}),
while expertise-aware prompting uses the same structured assessment
template as our policy model (Figure~\ref{fig:policy-prompt}) but
without RL training. For reviewer retrieval, we use a zero-shot scoring
template that elicits a 1--5 expertise rating for pairwise ranking
(Figure~\ref{fig:LLM-based-prompt}).

\section{Example Expertise Rubric}
\label{app:human-written-rubric}

Table~\ref{tab:rate-rubric-demo} presents a human-written example of a
paper-specific expertise rubric. This exemplar is provided to the LLM
as an in-context demonstration during rubric generation (see the prompt
in Figure~\ref{fig:rubric-generation-prompt}). It illustrates the
target structure: 1--2 Core criteria (weight~5) capturing expertise
essential to the paper's central contribution, and several Secondary
criteria (weight~3--4) covering supporting knowledge. Criterion
descriptions specify reviewer capabilities rather than paper-specific
details, ensuring that the rubric generalizes to candidate reviewers
with diverse publication backgrounds.

\section{Case Studies} 
\label{app:case-study}

We present case studies examining three aspects of our framework: the MERIT-Assessor's prediction behavior (Section~\ref{app:case-study-assessor}),
errors corrected by the MERIT-Retriever over the strongest baseline (Section~\ref{app:case-study-retriever}), and the retriever's remaining failure modes (Section~\ref{app:case-study-failure}).

\subsection{Assessor Case Studies}
\label{app:case-study-assessor}

Tables~\ref{tab:case-study-positive}--\ref{tab:case-study-false-negative} cover the four prediction outcomes of the MERIT-Assessor. 
The correct predictions illustrate how the model grounds suitability decisions in paper-specific expertise requirements and evidence from candidate
reviewers' prior work. 
The error cases reveal a recurring pattern in
evidence-transfer calibration: the model may over-transfer superficially related terminology across methodological contexts (false positive) or
under-transfer relevant expertise when the candidate's work differs in a specific technical variant (false negative).

\subsection{Retriever Error Corrections}
\label{app:case-study-retriever}

To understand what types of errors MERIT-Retriever corrects, we analyze LR-Bench preference pairs where RATE ranks incorrectly but MERIT ranks correctly.
Among the 88 corrected pairs, 63 (71.6\%) have an
rating gap of~1, compared with 19 (21.6\%) for gap~2 and 6 (6.8\%) for gap~3. 
Among the errors corrected by MERIT, most involve
adjacent expertise ratings, indicating that MERIT can resolve many close-call distinctions that RATE misses.
Two representative cases are presented below.

\paragraph{Case 1: Anchor paper.} \textit{HASH-RAG: Bridging Deep Hashing with Retriever for Efficient, Fine Retrieval and Augmented Generation}

\paragraph{Preference and model scores.}
\begin{center}
\begin{tabular}{lcc}
\toprule
 & Positive & Negative \\
\midrule
Expert preference & 5 & 4 \\
RATE & 0.7266 & 0.7364 \\
MERIT & 0.7627 & 0.6663 \\
\bottomrule
\end{tabular}
\end{center}

\paragraph{Interpretation.}
Although the anchor paper contains strong RAG-related surface terms, its core contribution is a hashing-based retrieval framework that learns binary hash codes for efficient storage and fine-grained retrieval.
The positive reviewer has direct expertise in semantic hashing and hash-code learning, while the negative reviewer is primarily oriented toward general RAG and LLM-based retrieval. 
RATE ranks the negative reviewer higher, possibly due to broader retrieval-related topical overlap.
MERIT instead gives greater weight to the paper's methodological core---hashing-based retrieval---suggesting that suitability supervision derived from criterion-level assessment helps the retriever distinguish central expertise from broader topical overlap.

\paragraph{Case 2: Anchor paper.} \textit{DVLTA-VQA: Decoupled Vision-Language Modeling with Text-Guided Adaptation for Blind Video Quality Assessment}

\paragraph{Preference and model scores.}
\begin{center}
\begin{tabular}{lcc}
\toprule
 & Positive & Negative \\
\midrule
Expert preference & 5 & 4 \\
RATE & 0.7091 & 0.7140 \\
MERIT & 0.8468 & 0.7127 \\
\bottomrule
\end{tabular}
\end{center}

\paragraph{Interpretation.}
The anchor paper is broadly related to visual-quality modeling, but its central evaluation methodology requires expertise in video and image quality assessment together with VLM/CLIP-based quality modeling, rather than general image aesthetics or video restoration. 
The positive reviewer has published directly on CLIP-based video quality assessment, while the negative reviewer focuses on image aesthetics and video super-resolution. 
RATE captures broad visual-quality similarity but does
not distinguish these roles; MERIT identifies the expertise more directly relevant to the paper's central methodology.

These cases suggest that MERIT better distinguishes expertise aligned with a paper's central contribution from broader topical overlap.

\subsection{Retriever Failure Modes}
\label{app:case-study-failure}

We inspect MERIT-Retriever's misranked LR-Bench pairs and find that remaining errors often occur when both candidates are strongly related to the anchor and the correct ordering depends on finer distinctions in
expertise breadth, specialization, or strength.
Two representative cases below each illustrate a different failure subtype.

\paragraph{Case 1: Anchor reviewer.}
\textit{Generative Large Recommendation Models: Emerging Trends in LLMs
for Recommendation}; \textit{A Unified Framework for Adaptive
Representation Enhancement and Inversed Learning in Cross-Domain
Recommendation}; \textit{Multi-granularity Interest Retrieval and
Refinement Network for Long-Term User Behavior Modeling in CTR
Prediction}.

\paragraph{Preference and model scores.}
\begin{center}
\begin{tabular}{lcc}
\toprule
 & Positive & Negative \\
\midrule
Expert preference & 5 & 4 \\
MERIT & 0.8556 & 0.8603 \\
\bottomrule
\end{tabular}
\end{center}

\paragraph{Interpretation.}
The reviewer's profile spans multiple recommendation sub-areas, including LLM-based recommendation, cross-domain recommendation, and CTR prediction.
The positive paper is a broad survey on data scarcity in
recommendation systems, whose evaluation demands breadth across the field.
The negative paper targets long sequence modeling in industrial recommenders, offering a more concentrated technique-level connection to the reviewer's CTR work. MERIT assigns high scores to both but slightly
prefers the technique-level match, reversing the expert's preference for the survey.
This suggests a breadth--depth calibration gap: MERIT may
over-emphasize a concentrated methodological match when breadth across the field is itself the relevant expertise requirement.

\paragraph{Case 2: Anchor reviewer.}
\textit{Federated Markov Imputation: Privacy-Preserving Temporal
Imputation in Multi-Centric ICU Environments};
\textit{Distribution-Controlled Client Selection to Improve Federated
Learning Strategies}; \textit{Improving Early Sepsis Onset Prediction
Through Federated Learning}.

\paragraph{Preference and model scores.}
\begin{center}
\begin{tabular}{lcc}
\toprule
 & Positive & Negative \\
\midrule
Expert preference & 4 & 3 \\
MERIT & 0.7994 & 0.8136 \\
\bottomrule
\end{tabular}
\end{center}

\paragraph{Interpretation.}
The reviewer has direct prior work on client selection in federated learning.
The positive paper proposes a Shapley-based client selection method, matching this specialization directly, while the negative paper addresses federated-learning heterogeneity more generally through similarity and complementarity. 
MERIT captures the shared federated-learning domain but fails to sufficiently prioritize the reviewer's more specific specialization in client selection.

These cases highlight a limitation of binary suitability supervision: it provides limited signal for finer distinctions in expertise breadth, specialization, and strength. More fine-grained preference supervision may help resolve such cases.

\begin{table*}[t] 
\label{tab:rate-rubric-demo}
\centering
\small
\setlength{\tabcolsep}{6pt}
\renewcommand{\arraystretch}{1.12}

\caption{Human-written expertise rubric for the paper \emph{Search-R1}.
Each row reports a criterion title $t_j$ and description $d_j$,
together with its Core/Secondary role and importance weight $w_j$.}
\label{tab:rate-rubric-demo}

\begin{tabularx}{\textwidth}{
  @{}
  >{\raggedright\arraybackslash}p{0.25\textwidth}
  >{\centering\arraybackslash}p{0.14\textwidth}
  >{\raggedright\arraybackslash}X
  @{}
}
\toprule
\multicolumn{3}{@{}l}{\textbf{Input paper}} \\
\multicolumn{3}{@{}p{\textwidth}@{}}{
\textit{Search-R1: Training LLMs to Reason and Leverage Search Engines with Reinforcement Learning}
} \\
\midrule
\textbf{Criterion} & \textbf{Type} & \textbf{Description} \\
\midrule

RLHF \& RL for LLMs
&
\begin{tabular}[t]{@{}c@{}}
\textbf{Core}\\[-1pt]
\textit{Weight 5}
\end{tabular}
&
Strong expertise in applying reinforcement learning algorithms (specifically PPO or GRPO) to large language models, including understanding policy optimization stability and reward function design.
\\
\addlinespace[5pt]

Retrieval-Augmented Generation (RAG)
&
\begin{tabular}[t]{@{}c@{}}
\textbf{Core}\\[-1pt]
\textit{Weight 5}
\end{tabular}
&
Deep understanding of RAG architectures, particularly multi-turn retrieval, query generation strategies, and the integration of external search results into LLM context.
\\
\addlinespace[5pt]

Agentic Tool Use \& Environments
&
\begin{tabular}[t]{@{}c@{}}
\textbf{Secondary}\\[-1pt]
\textit{Weight 4}
\end{tabular}
&
Familiarity with modeling LLMs as agents that interact with external environments (search engines) via special tokens or API calls in an interleaved manner.
\\
\addlinespace[5pt]

Reasoning \& Chain-of-Thought
&
\begin{tabular}[t]{@{}c@{}}
\textbf{Secondary}\\[-1pt]
\textit{Weight 3}
\end{tabular}
&
Knowledge of advanced reasoning frameworks (e.g., Chain-of-Thought, DeepSeek-R1 styles) to evaluate the quality of the model's generated reasoning trajectories.
\\

\bottomrule
\end{tabularx}
\end{table*}
\definecolor{CaseNavy}{HTML}{2E4057}
\definecolor{CaseRed}{HTML}{8B2C2C}
\definecolor{CoreNavy}{HTML}{3E5C76}
\definecolor{SecSlate}{HTML}{6C7A89}
\definecolor{DecisionSage}{HTML}{5B7C69}
\definecolor{DecisionBurgundy}{HTML}{7A4E57}

\definecolor{BgBlue}{HTML}{F6F8FB}
\definecolor{BgGray}{HTML}{F7F7F8}
\definecolor{BgGreen}{HTML}{F4F8F4}
\definecolor{BgRose}{HTML}{FAF5F6}

\definecolor{BgTakeaway}{HTML}{F5F1F8}
\definecolor{TakeawayBrown}{HTML}{7A4E00}
\newcommand{\coretag}{%
  \begingroup
  \setlength{\fboxsep}{1.5pt}%
  \fcolorbox{CoreNavy}{CoreNavy!8}{\textcolor{CoreNavy}{\scriptsize\textsc{Core}}}%
  \endgroup
}

\newcommand{\sectag}{%
  \begingroup
  \setlength{\fboxsep}{1.5pt}%
  \fcolorbox{SecSlate}{SecSlate!8}{\textcolor{SecSlate}{\scriptsize\textsc{Secondary}}}%
  \endgroup
}

\newcommand{\decisiontag}{%
  \begingroup
  \setlength{\fboxsep}{1.5pt}%
  \fcolorbox{DecisionSage}{DecisionSage!8}{\textcolor{DecisionSage}{\scriptsize\textsc{Suitable}}}%
  \endgroup
}

\newcommand{\unsuitabletag}{%
  \begingroup
  \setlength{\fboxsep}{1.5pt}%
  \fcolorbox{DecisionBurgundy}{DecisionBurgundy!8}{\textcolor{DecisionBurgundy}{\scriptsize\textsc{Unsuitable}}}%
  \endgroup
}

\begin{table*}[t]
\centering
\small
\setlength{\tabcolsep}{7pt}
\renewcommand{\arraystretch}{1.16}

\caption{True positive case. The model correctly identifies
the reviewer as suitable based on strong evidence for core GEC
evaluation expertise.}
\label{tab:case-study-positive}

\begin{tabularx}{\textwidth}{
>{\raggedright\arraybackslash}X
}
\toprule
\textcolor{CaseNavy}{\textbf{Case study: Correct positive prediction}}
\\
\midrule

\rowcolor{BgBlue}
\textbf{Input: Target paper}

\vspace{0.25em}
\textbf{Title:} 
\textit{CLEME2.0: Towards Interpretable Evaluation by Disentangling Edits for Grammatical Error Correction}

\vspace{0.25em}
\textbf{TL;DR:}
The paper proposes an \textbf{interpretable evaluation metric} for grammatical error correction by disentangling edits into \textbf{fine-grained correction types} and evaluating consistency with \textbf{human judgments}.
\\

\midrule
\rowcolor{BgGray}
\textbf{Key fields from the policy-model output}
\\

\rowcolor{BgGray}
\textbf{[REQUIRED\_EXPERTISE]}

\vspace{0.18em}
\coretag \quad \textbf{Grammatical Error Correction evaluation metrics}

\coretag \quad \textbf{Edit-level analysis and metric interpretability}

\vspace{0.12em}
\sectag \quad \textbf{Reference-based vs. reference-free evaluation methods}

\vspace{0.12em}
\sectag \quad \textbf{Adversarial testing, robustness, and human consistency of GEC metrics}
\\

\addlinespace[0.35em]

\rowcolor{BgGray}
\textbf{[EVIDENCE\_CLUES]}

\vspace{0.18em}
\textit{Rethinking Evaluation Metrics for Grammatical Error Correction}:
Supports \textbf{CORE} (GEC evaluation), \textbf{Secondary} (human-aligned evaluation), and \textbf{Secondary} (benchmarking of automatic metrics).

\vspace{0.18em}
\textit{Reliability Crisis of Reference-free Metrics}:
Supports \textbf{Secondary} (adversarial testing), \textbf{Secondary} (reference-free metrics), and \textbf{Secondary} (robustness).

\vspace{0.18em}
\textit{Improving Explainability of Sentence-level Metrics}:
Supports \textbf{Secondary} (attribution methods) and \textbf{Secondary} (interpretability).
\\

\addlinespace[0.35em]

\rowcolor{BgGray}
\textbf{[REASONING\_SUMMARY]}

\vspace{0.18em}
The candidate has extensive expertise in GEC evaluation, including both reference-based and reference-free metrics, and their limitations.

\vspace{0.18em}
Their work on human consistency, benchmarking, and meta-evaluation aligns with the target paper's evaluation on human-judgment datasets and multiple reference datasets.

\vspace{0.18em}
Their work on adversarial attacks further provides supporting expertise on the reliability limitations of automatic GEC metrics.
\\

\midrule
\rowcolor{BgGreen}
\textbf{Policy-model decision}

\vspace{0.25em}
\decisiontag
\quad
\textbf{Final label: 1.}
\\

\midrule
\rowcolor{BgTakeaway}
\textbf{Interpretation.}

\vspace{0.18em}
This case illustrates how the policy model identifies paper-specific expertise requirements, links them to evidence from a reviewer's prior work, and produces an evidence-grounded positive suitability decision.
\\
\bottomrule
\end{tabularx}
\end{table*}


\begin{table*}[t]
\centering
\small
\setlength{\tabcolsep}{7pt}
\renewcommand{\arraystretch}{1.16}

\caption{True negative case. The model correctly identifies partial
topical overlap in graph-based recommendation but insufficient evidence
for the paper's core requirements in rule-driven attribute embedding.}
\label{tab:case-study-negative}

\begin{tabularx}{\textwidth}{
>{\raggedright\arraybackslash}X
}
\toprule
\textcolor{CaseNavy}{\textbf{Case study: Correct negative prediction}}
\\
\midrule

\rowcolor{BgBlue}
\textbf{Input: Target paper}

\vspace{0.25em}
\textbf{Title:}
\textit{RAE: A Rule-Driven Approach for Attribute Embedding in Property Graph Recommendation}

\vspace{0.25em}
\textbf{TL;DR:}
The paper proposes a \textbf{rule-driven attribute embedding} framework for property graph recommendation. 
It mines \textbf{semantic dependency rules} from property graphs, uses them to guide random walks for enriched attribute embeddings, and integrates these embeddings into GCN-based recommendation models to improve robustness under \textbf{data sparsity} and \textbf{attribute missingness}.
\\

\midrule
\rowcolor{BgGray}
\textbf{Key fields from the policy-model output}
\\

\rowcolor{BgGray}
\textbf{[REQUIRED\_EXPERTISE]}

\vspace{0.18em}
\coretag \quad \textbf{Rule-driven attribute embedding in property graphs}

\vspace{0.18em}
\coretag \quad \textbf{Semantic rule mining and attribute dependency modeling}

\vspace{0.18em}
\sectag \quad \textbf{GNN/GCN-based recommendation}

\vspace{0.18em}
\sectag \quad \textbf{Data sparsity and attribute missingness}
\\

\addlinespace[0.35em]

\rowcolor{BgGray}
\textbf{[EVIDENCE\_CLUES]}

\vspace{0.18em}
\textit{Federated Heterogeneous Graph Neural Network for Privacy-preserving Recommendation}:
Provides partial overlap in \textbf{HIN-based recommendation} and \textbf{graph neural networks}, but does not address \textbf{rule-driven attribute embedding}, \textbf{property-graph attribute dependencies}, or \textbf{semantic rule mining}.

\vspace{0.18em}
\textit{Federated Graph Condensation with Information Bottleneck Principles}:
Focuses on \textbf{graph condensation}, \textbf{federated graph learning}, and \textbf{privacy protection}, rather than \textbf{attribute embeddings} or \textbf{rule-guided representation learning}.

\vspace{0.18em}
\textit{Phantom Subgroup Poisoning: Stealth Attacks on Federated Recommender Systems}:
Studies \textbf{poisoning attacks} in federated recommender systems, but not \textbf{attribute-based recommendation}, \textbf{property graphs}, or \textbf{semantic dependency rules}.

\vspace{0.18em}
\textit{Rethinking Byzantine Robustness in Federated Recommendation}:
Focuses on \textbf{Byzantine robustness} and \textbf{sparse aggregation} in federated recommendation, rather than \textbf{attribute modeling} or \textbf{rule-based embeddings}.
\\

\addlinespace[0.35em]

\rowcolor{BgGray}
\textbf{[REASONING\_SUMMARY]}

\vspace{0.18em}
The candidate has related expertise in \textbf{graph-based recommendation}, \textbf{federated learning}, and \textbf{robustness/security} for recommender systems.

\vspace{0.18em}
However, the target paper's main contribution is not federated recommendation or graph robustness, but a \textbf{rule-driven attribute embedding} method that mines \textbf{semantic dependency rules} from property graphs and uses them to guide representation learning.

\vspace{0.18em}
The candidate's prior work provides only partial overlap through HINs, GNNs, and sparsity-aware recommendation, but does not provide direct evidence of expertise in \textbf{property-graph attribute modeling}, \textbf{semantic rule mining}, or \textbf{rule-guided attribute embeddings}.
\\

\midrule
\rowcolor{BgGreen}
\textbf{Policy-model decision}

\vspace{0.25em}
\unsuitabletag
\quad
\textbf{Final label: 0.} 
\\

\midrule
\rowcolor{BgTakeaway}
\textbf{Interpretation.}

\vspace{0.12em}
This case illustrates how the policy model distinguishes partial topical overlap from true reviewer suitability: although the candidate has related expertise in graph and federated recommendation, the model correctly identifies the lack of evidence for the target paper's core requirements in rule-driven attribute embedding and semantic rule mining.
\\
\bottomrule
\end{tabularx}
\end{table*}

\begin{table*}[t]
\centering
\small
\setlength{\tabcolsep}{7pt}
\renewcommand{\arraystretch}{1.16}

\caption{False positive case. The model incorrectly transfers
superficially related terminology (counterfactual explanations,
predictive multiplicity) to the target paper's causal inference
requirements.}
\label{tab:case-study-false-positive}

\begin{tabularx}{\textwidth}{
>{\raggedright\arraybackslash}X
}
\toprule
\textcolor{CaseRed}{\textbf{Case study: False positive prediction}}
\\
\midrule

\rowcolor{BgBlue}
\textbf{Input: Target paper}

\vspace{0.25em}
\textbf{Title:}
\textit{Self Balancing Neural Network: A Novel Method to Estimate Average Treatment Effect}

\vspace{0.25em}
\textbf{TL;DR:}
The paper proposes a \textbf{self-balancing neural network} for estimating the \textbf{average treatment effect} in observational studies. 
It addresses bias from \textbf{confounding variables} and \textbf{propensity score misspecification} by learning pseudo propensity scores through a balancing network integrated into the outcome model.
\\

\midrule
\rowcolor{BgGray}
\textbf{Key fields from the policy-model output}
\\

\rowcolor{BgGray}
\textbf{[REQUIRED\_EXPERTISE]}

\vspace{0.18em}
\coretag \quad \textbf{Causal inference in observational studies}

\vspace{0.18em}
\coretag \quad \textbf{Propensity score modeling and balancing techniques}

\vspace{0.18em}
\sectag \quad \textbf{Machine learning methods for causal estimation}

\vspace{0.18em}
\sectag \quad \textbf{Addressing model misspecification in causal models}

\vspace{0.18em}
\sectag \quad \textbf{Neural network architectures for causal inference}
\\

\addlinespace[0.35em]

\rowcolor{BgGray}
\textbf{[EVIDENCE\_CLUES]}

\vspace{0.18em}
\textit{Explainable Bank Failure Prediction Models}:
Discusses \textbf{counterfactual explanations}, \textbf{model interpretability}, and the reliability of black-box predictive models.

\vspace{0.18em}
\textit{Predictive Multiplicity in Survival Models}:
Addresses \textbf{predictive multiplicity}, model disagreement, and uncertainty in survival-based predictive modeling.

\vspace{0.18em}
\textit{Beyond the Single-Best Model: Rashomon Partial Dependence Profile}:
Explores \textbf{model multiplicity}, \textbf{explanation uncertainty}, and trustworthy model interpretation.
\\

\addlinespace[0.35em]

\rowcolor{BgGray}
\textbf{[REASONING\_SUMMARY]}

\vspace{0.18em}
The candidate's work spans \textbf{causal inference}, \textbf{model reliability}, and \textbf{predictive multiplicity}, which are central to the target paper's focus on addressing confounding and model misspecification.

\vspace{0.18em}
While the candidate does not explicitly mention \textbf{self-balancing neural networks} or \textbf{propensity score modeling}, their focus on model consistency, uncertainty, and robustness aligns with the paper's technical contributions.
\\

\midrule
\rowcolor{BgRose}
\textbf{Policy-model decision}

\vspace{0.25em}
\decisiontag
\quad
\textbf{Predicted label: 1.}
\\

\midrule
\rowcolor{BgTakeaway}
\textbf{Interpretation.}

\vspace{0.12em}
This error case reveals a \textbf{terminology-transfer error}: the policy model transfers superficially related terms across different methodological contexts. 
In particular, \textbf{counterfactual explanations} in explainable AI do not provide direct evidence of expertise in observational causal inference, and \textbf{predictive multiplicity} or generic model reliability does not establish expertise in \textbf{propensity score modeling} or \textbf{average treatment effect estimation}.
\\

\bottomrule
\end{tabularx}
\end{table*}


\begin{table*}[t]
\centering
\small
\setlength{\tabcolsep}{7pt}
\renewcommand{\arraystretch}{1.16}

\caption{False negative case. The model applies overly strict
technique-specific matching, underestimating the reviewer's transferable
expertise in diffusion-based image editing.}
\label{tab:case-study-false-negative}

\begin{tabularx}{\textwidth}{
>{\raggedright\arraybackslash}X
}
\toprule
\textcolor{CaseRed}{\textbf{Case study: False negative prediction}}
\\
\midrule

\rowcolor{BgBlue}
\textbf{Input: Target paper}

\vspace{0.25em}
\textbf{Title:}
\textit{FramePainter: Endowing Interactive Image Editing with Video Diffusion Priors}

\vspace{0.25em}
\textbf{TL;DR:}
The paper proposes \textbf{FramePainter}, an interactive image-editing method that reformulates image editing as an \textbf{image-to-video generation} task. 
It leverages \textbf{video diffusion priors} from Stable Video Diffusion and introduces \textbf{matching attention} to improve dense correspondence and visual consistency under large object motion.
\\

\midrule
\rowcolor{BgGray}
\textbf{Key fields from the policy-model output}
\\

\rowcolor{BgGray}
\textbf{[REQUIRED\_EXPERTISE]}

\vspace{0.18em}
\coretag \quad \textbf{Video diffusion models for image editing}

\vspace{0.18em}
\sectag \quad \textbf{Techniques for handling dynamic motion between frames}

\vspace{0.18em}
\sectag \quad \textbf{Attention mechanisms for image-to-video generation}

\vspace{0.18em}
\sectag \quad \textbf{Consistent object editing with diffusion models}

\vspace{0.18em}
\sectag \quad \textbf{Image editing with spatial-temporal correspondence}

\vspace{0.18em}
\sectag \quad \textbf{Evaluation of generalization in out-of-domain scenarios}
\\

\addlinespace[0.35em]

\rowcolor{BgGray}
\textbf{[EVIDENCE\_CLUES]}

\vspace{0.18em}
\textit{PixelMan: Consistent Object Editing with Diffusion Models via Pixel Manipulation and Generation}:
Supports \textbf{Secondary} (consistent object editing with diffusion models).

\vspace{0.18em}
\textit{GOAt: Explaining Graph Neural Networks via Graph Output Attribution},
\textit{Building Optimal Neural Architectures using Interpretable Knowledge}, and
\textit{EiG-Search: Generating Edge-Induced Subgraphs for GNN Explanation in Linear Time}:
Treated as irrelevant to the target paper.
\\

\addlinespace[0.35em]

\rowcolor{BgGray}
\textbf{[REASONING\_SUMMARY]}

\vspace{0.18em}
The candidate's work is focused on \textbf{graph neural networks}, \textbf{neural architecture search}, and \textbf{diffusion-based object editing}.

\vspace{0.18em}
While \textit{PixelMan} involves \textbf{diffusion models for object editing}, it is limited to static image manipulation and does not address \textbf{video diffusion priors}, \textbf{dynamic motion handling}, or \textbf{image-to-video generation}.

\vspace{0.18em}
The candidate lacks explicit expertise in \textbf{video diffusion models}, \textbf{attention mechanisms for temporal correspondence}, and the specific technical challenges of \textbf{FramePainter}, such as leveraging video diffusion priors for interactive editing and handling large motion between frames.
\\

\midrule
\rowcolor{BgRose}
\textbf{Policy-model decision}

\vspace{0.25em}
\unsuitabletag
\quad
\textbf{Predicted label: 0.}
\\

\midrule
\rowcolor{BgTakeaway}
\textbf{Interpretation.}

\vspace{0.12em}
This false negative reflects \textbf{overly strict technique-specific matching}: the policy model emphasizes the absence of direct evidence in \textbf{video diffusion priors}, \textbf{image-to-video generation}, and \textbf{matching attention}, but underestimates the candidate's transferable expertise in \textbf{diffusion-based image editing}, \textbf{consistent object manipulation}, and \textbf{appearance-preserving editing}.
\\

\bottomrule
\end{tabularx}
\end{table*}


\definecolor{promptbg}{HTML}{FAFBFC}
\definecolor{promptframe}{HTML}{D6DCE5}
\definecolor{prompttitlebg}{HTML}{EEF2F7}
\definecolor{prompttitlefg}{HTML}{1F2937}

\newtcblisting{promptbox}[1]{
  enhanced,
  listing only,
  breakable,
  width=\textwidth,
  colback=promptbg,
  colframe=promptframe,
  coltitle=prompttitlefg,
  colbacktitle=prompttitlebg,
  boxrule=0.6pt,
  arc=1.8mm,
  left=1.5mm,
  right=1.5mm,
  top=1.2mm,
  bottom=1.2mm,
  title={#1},
  fonttitle=\bfseries\small,
  attach boxed title to top left={xshift=2mm,yshift*=-2mm},
  boxed title style={
    sharp corners,
    boxrule=0.5pt,
    arc=1mm
  },
  listing options={
    basicstyle=\ttfamily\scriptsize,
    breaklines=true,
    breakatwhitespace=false,
    columns=fullflexible,
    keepspaces=true,
    showstringspaces=false
  }
}

\begin{figure*}[t]
\caption{Prompt template for paper-specific expertise rubric generation.}
\centering
\begin{promptbox}{}
### System Prompt
You generate reviewer expertise rubrics for a research paper.

Task:
Infer the generalized expertise a qualified reviewer should have based on the paper title, abstract, and introduction.

Rules:
- Do not require expertise in the paper's specific proposed method, acronym, dataset, or baseline names.
- Abstract paper-specific details into transferable expertise such as domain knowledge, methodology, theory, or evaluation skills.
- Keep rubric items distinct and avoid overlap.
- Prefer the paper's core methodological or theoretical expertise over broad topic labels.
- Use the smallest sufficient set of expertise areas needed to evaluate the paper.
- Do not split closely related skills into separate items.
- Add an item only if it captures a distinct reviewer expertise that cannot be merged with another item.

Output:
Return only a JSON array of 3 to 6 items.
Use 1 or 2 CORE items only.
Use 2 CORE items only when the paper requires two distinct and indispensable expertise areas; otherwise use 1.
All remaining items must be Secondary.

Each item must have exactly:
- "title"
- "description"
- "weight"

Description prefix rules:
- If weight = 5, the description must begin with "CORE:".
- If weight = 3 or 4, the description must begin with "Secondary:".

Weight meaning:
- CORE Criteria (Weight: 5)
  - Definition: The indispensable expertise required to understand and evaluate the paper's main contribution.
- Secondary Criteria (Weight: 3-4)
  - Definition: Supporting expertise needed to verify important methodological, theoretical, empirical, or evaluation details.

Quality requirements:
- Titles should describe reviewer expertise, not paper modules.
- Descriptions should specify the scope of the required expertise and explain how it is needed for evaluating the paper.
- Avoid overly fine-grained or redundant criteria.
- Prefer concise descriptions.

### Examples
{examples_string}

------------------------------------------------------------

### User Prompt
[Paper Title]
{paper_title}

[Paper Abstract]
{paper_abstract}

[Paper Introduction]
{paper_introduction}

Generate the reviewer-expertise rubric as a JSON array.
Return JSON only.
\end{promptbox}

\label{fig:rubric-generation-prompt}
\end{figure*}

\begin{figure*}[t]
\caption{Prompt template for the policy model's reviewer suitability
assessment. This template is also used by the expertise-aware prompting
baselines.}
\centering
\begin{promptbox}{}
### System Prompt
You are an expert Conference Area Chair. Your task is to make a final "Accept/Reject" decision on whether a Candidate Reviewer is qualified to review a specific Target Paper.

Input Data:
1. Target Paper (Title, Abstract, Introduction)
2. Candidate Author Profile (List of historical publications)

Decision Logic:
- Label 1 (Qualified): The candidate possesses **at least ONE of the CORE (Primary)** expertise requirements **AND** at least one additional relevant skill (Secondary or another Core). They have sufficient domain knowledge to evaluate the paper's key contributions.
- Label 0 (Unqualified): The candidate matches ONLY secondary aspects, OR matches a CORE concept in isolation without any supporting context (e.g., knows the general theory but lacks the specific application domain), OR has no relevance at all.

Instruction:
You must strictly follow the output structure below. 

Step 1: Analyze the Target Paper. Extract **3 to 6** distinct expertise requirements.
   - **Constraint:** Select the **most critical** topics. Do not list more than 6 items to avoid dilution.
   - Mark the central problem/method as "(CORE)" and enabling technologies as "(Secondary)".
   - **Important:** Define CORE as the **underlying problem/method** (e.g., "Diffusion Models") rather than the specific novelty.
Step 2: Scan the Candidate's profile. Select specific papers that serve as evidence. 
   - **Crucial:** Look for **semantic matches**.
Step 3: Synthesize the analysis. Evaluate the match based on the **Combination Rule**.
   - **Threshold:** The candidate MUST demonstrate expertise in **at least ONE CORE requirement**, PLUS **at least one additional requirement** (which can be a Secondary or another CORE). 
   - Note: Do NOT demand the candidate to master ALL COREs if multiple exist, as interdisciplinary papers often require a panel of different experts.

Output Format (Strictly use these headers):
[REQUIRED_EXPERTISE]
<List **3 to 6** requirements here, one per line. Be granular but focus on capabilities. Example:
1. Adversarial Patch Attacks on Vision Transformers (CORE)
2. Quantization-Aware Training (Secondary)
3. Image Classification Benchmarks (Secondary)>

[EVIDENCE_CLUES]
<List specific paper titles and a brief tag indicating which requirement they support. Example:
- "DeepPatch: Attacking ViTs": Supports CORE (Adversarial Patch Attacks).
- "Q-ViT: Robust Quantization": Supports Secondary (Quantization).
If no relevant papers are found, write 'None'.>

[REASONING_SUMMARY]
<Provide a holistic analysis. Explicitly discuss:
1. Whether the candidate covers the (CORE) expertise.
2. Which (additional) expertise supports the CORE match.
3. Conclusion based on the "Core + 1 additional" rule.>

[FINAL_LABEL]
<0 or 1>

------------------------------------------------------------

### User Prompt
[[Target Paper]]
Title: {{paper_title}}
Content: {{paper_abstract}} 
{{paper_introduction}}

[[Candidate Author Profile]]
(Recent Publications)
{{candidate_history_list}}

Based on the strict standard (1=Expert Match, 0=Everything Else), evaluate this pair.
\end{promptbox}

\label{fig:policy-prompt}
\end{figure*}

\begin{figure*}[t]
\caption{Prompt template for the LLM judge used to compute
rubric-guided gated rewards.}
\centering
\begin{promptbox}{}
### System Prompt
You are the Senior Program Chair (SAC) for a top-tier Computer Science conference.
You are auditing a "Reviewer Suitability Analysis" to ensure the evaluation is fair, evidence-based, and logically sound.

Your Goal: Detect incompetent analysis, logical contradictions, or "gaming the system" (e.g., keyword stuffing, rejecting qualified candidates to play it safe, or accepting unqualified ones).

You have access to:
1. Target Paper Context.
2. Candidate's Verified Publications.
3. Gold Rubrics.
4. The Report to be audited.

Output Structure:
Output **ONLY** the strict JSON object.

JSON Schema:
1. "rubric_breakdown": { "Rubric Title": boolean }
   - true: The Report **actively addresses and discusses** this requirement. 
     * It goes beyond merely listing the string; it evaluates whether the candidate possesses or lacks this specific expertise (e.g., linking it to a specific paper or explicitly stating it is missing in the reasoning).
   - false: The requirement is ignored, OR it is merely "keyword stuffed" (copy-pasted into a list without being analyzed against the candidate's profile).

2. "logical": Boolean.
   - true: The decision is **sound, fair, and strongly supported** by the "1 Core + 1 additional requirement" threshold rule.
     * If Label 1: The Report accurately identifies at least ONE Core and ONE additional requirement, supported by valid cited papers.
     * If Label 0: The Report correctly proves the candidate lacks Core expertise or lacks additional requirement support.
   - false: The Analysis contains a **logical failure, contradiction, or unfairness**. Examples of logical failures:
     * **False Positive (Bad 1):** Giving Label 1 based on weak/tangential evidence (e.g., Candidate knows broad "AI", Paper needs "Diffusion" -> logical is false) or hallucinated evidence.
     * **False Negative (Bad 0):** Giving Label 0 despite the candidate clearly meeting the "1 Core + 1 additional" threshold (e.g., demanding a 100\% perfect match on a specific novelty when the candidate has sufficient foundational expertise).
     * **Contradiction:** The text argues the candidate is a strong match but outputs Label 0 (or vice versa).

------------------------------------------------------------

### User Prompt
=== 1. Target Paper Context ===
Title: 
{{paper_title}}

Abstract: 
{{paper_abstract}}

Introduction: 
{{paper_introduction}}

=== 2. Candidate Profile ===
{{candidate_history_list}}

=== 3. Gold Rubrics ===
{{rubrics_json}}

=== 4. Report (To be Audited) ===
{{actor_output_text}}

=== Task ===
Audit the Report fairly and objectively. 
1. **Check for Keyword Stuffing:** Mark rubric items as false if they are just listed in the requirements section but never actually evaluated against the candidate's history in the reasoning/evidence.
2. **Check for Logical Consistency:** Ensure the final label accurately reflects the evidence. Penalize over-claiming (unjustified 1s) AND overly strict rejections (unjustified 0s when the candidate meets the 1 Core + 1 additional requirement threshold).

Generate the strict JSON output:
1. "rubric_breakdown": { "Rubric Title": boolean } 
2. "logical": boolean.
\end{promptbox}

\label{fig:judge-prompt}
\end{figure*}

\begin{figure*}[t]
\caption{Prompt template for the direct prompting baseline in reviewer
suitability classification.}
\centering
\begin{promptbox}{}
### System Prompt
You are an expert Conference Area Chair. Determine whether a Candidate Reviewer is qualified to review a Target Paper.

Input:
1. Target Paper (Title, Abstract, Introduction)
2. Candidate Author Profile (historical publications)

Decision Rule:
- Label 1: The candidate has sufficient expertise to review the target paper.
- Label 0: The candidate lacks the expertise needed to provide a competent review.

Output Format:
[FINAL_LABEL]
<0 or 1>
------------------------------------------------------------

### User Prompt
[[Target Paper]]
Title: {{paper_title}}
Content: {{paper_abstract}} 
{{paper_introduction}}

[[Candidate Author Profile]]
(Recent Publications)
{{candidate_history_list}}

Based on the candidate's research background, are they a suitable reviewer for this target paper?
\end{promptbox}

\label{fig:direct-prompting-prompt}
\end{figure*}

\begin{figure*}[t]
\caption{Prompt template for zero-shot LLM scoring in reviewer
retrieval evaluation.}
\centering
\begin{promptbox}{}
Task: Score the reviewer expertise and fit for reviewing the target paper.

[Target Paper]
Title: {paper_title}
Abstract: {paper_abstract}

[Reviewer Profile]
Recent Publications: {reviewer_papers}

[Scoring Rubric]
Please choose one integer score from 1 to 5:

5 (Top Expert): Active researcher in this specific sub-field; published highly relevant work; capable of writing a similar paper.
4 (Expert): Very familiar with the general field; can reproduce methods and judge technical quality.
3 (Knowledgeable): Works in a related field; understands core concepts but no direct work in this sub-direction.
2 (Vague Familiarity): General awareness; can understand the abstract but is unfamiliar with technical nuances.
1 (No Expertise): No background; cannot understand terminology or core logic.

At the end of your response, you must output only one integer from 1 to 5, and nothing else.
\end{promptbox}

\label{fig:LLM-based-prompt}
\end{figure*}

\end{document}